\documentclass[runningheads]{llncs}

\usepackage[T1]{fontenc}
\usepackage{graphicx,verbatim}
\usepackage{booktabs} 
\usepackage{array,multirow,graphicx,stackengine}
\usepackage{color, colortbl}
\usepackage[dvipsnames]{xcolor}
\usepackage{caption}
\usepackage{tabularx,booktabs}
\newcolumntype{Y}{>{\centering\arraybackslash}X}
\usepackage{tikz}
\usepackage{bm}
\usepackage{enumitem}

\usepackage{amssymb}
\usepackage{pifont}

\usepackage{amsmath}
\usepackage{bbm}
\usepackage{yhmath}
\usepackage{amssymb}
\usepackage{mathtools}
\usepackage{upgreek}
\usepackage{amsthm}
\usepackage{cite}

\usepackage[colorlinks=true, urlcolor=blue, linkcolor=red]{hyperref}
\begin{document}
\title{Finer Disentanglement of Aleatoric Uncertainty Can Accelerate Chemical Histopathology Imaging}
%

\author{Ji-Hun Oh\inst{\dag}\and
Kianoush Falahkheirkhah\inst{\dag}\and
Rohit Bhargava\inst{\dag, \ddag}\\
}
%
\authorrunning{JH Oh et al.}

%
\institute{
\textsuperscript{\dag} University of Illinois Urbana-Champaign, Urbana, IL, USA\\
\textsuperscript{\ddag} CZ Biohub Chicago, LLC, Chicago, IL, USA\\
{\tt\small \{jihunoh2, kf4, rxb\}@illinois.edu}\\
}
    
\maketitle              
\begin{abstract}
Label-free chemical imaging holds significant promise for improving digital pathology workflows, but data acquisition speed remains a limiting factor. To address this gap, we propose an adaptive strategy—initially scan the low information (LI) content of the entire tissue quickly, identify regions with high aleatoric uncertainty (AU), and selectively re-image them at better quality to capture higher information (HI) details. The primary challenge lies in distinguishing between high-AU regions mitigable through HI imaging and those that are not. However, since existing uncertainty frameworks cannot separate such AU subcategories, we propose a fine-grained disentanglement method based on post-hoc latent space analysis to unmix resolvable from irresolvable high-AU regions. We apply our approach to streamline infrared spectroscopic imaging of breast tissues, achieving superior downstream segmentation performance. This marks the first study focused on fine-grained AU disentanglement within dynamic image spaces (LI-to-HI), with novel application to streamline histopathology. Code will be made public.

\end{abstract}

%
%
%
\section{Introduction}
\label{sec1}
Histopathology, the practice of diagnosing and treating cancer, relies on special dyes like hematoxylin and eosin to highlight tissue and cellular architecture. While this process is well-established, challenges remain—recurrent costs of labor and reagents, data quality issues from staining artifacts, lab variability, and tissue damage, and interpretation subjectivity. The desire to overcome these limitations has driven interest in label-free imaging \cite{huang2023rapid, mittal2018simultaneous, fereidouni2017microscopy, hoover2013advances, park2018quantitative}, which extracts intrinsic sample information without binding dyes. In particular, infrared (IR) chemical imaging leverages the fundamental vibrational spectroscopic fingerprints of biomolecules—including proteins, lipids, nucleic acids, and collagen—to quantify their spatial distributions. Pairing this rich molecular detail with deep neural net (DNN) tools like image segmenters enhances digital pathology, potentially offering scalability and robustness \cite{bhargava2023digital}.

\begin{figure}[t]
\centering
\includegraphics[width=\textwidth]{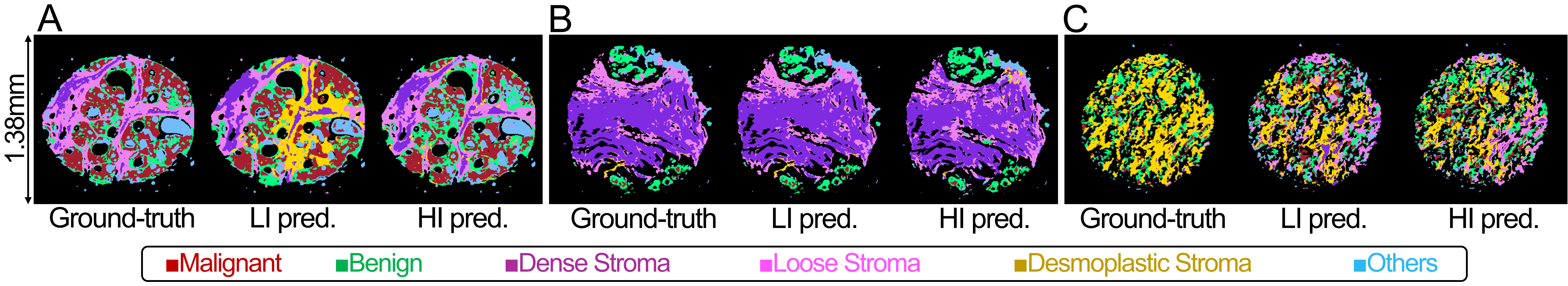}
\caption{\textbf{Impact of data quality.} We show segmentation results of LI and HI IR images of breast tissues (\S\ref{sec41}). We observe discrepant behavior—A is accurate only at HI, B is performant at both HI and LI, while C is subpar throughout.} \label{fig1}
\end{figure}

However, acquiring such images with high-information (HI) content can take hours depending on technology and tissue size, posing challenges in low-resource or time-sensitive environments. A practical workaround is to minimize the data collected, such as employing fewer spectral frequencies or sparse scanning. Unfortunately, such low-information (LI) images often induce ambiguity (\textit{i.e.}, aleatoric uncertainty, AU), degrading downstream DNN performance (sample A in Fig. \ref{fig1}). On the brighter side, not all data suffer from this issue, with LI being sufficient (B in Fig. \ref{fig1}). This observation suggests an adaptive strategy that balances throughput and accuracy: {Scan the entire sample at LI, identify critical regions, and then re-image these areas at HI.} The crux lies in selecting regions with high AU while avoiding conflation with epistemic uncertainty (EU)—e.g., when data is out-of-distribution (OOD). While recent studies have made strides in disentangling these two types of uncertainty \cite{depeweg2018decomposition, kendall2017uncertainties, gruber2023uncertainty, mukhoti2023deep, sun2023flagged}, there remains another issue: \textit{not all AU is resolvable at HI} (C in Fig. \ref{fig1}). This calls for a further partition of high-AU data into {resolvable} \textit{vs.} {irresolvable} cases, minimizing the unnecessary sampling of the latter. No current uncertainty framework achieves this, as they do not go beyond the EU-AU dichotomy.

{Our contribution} is two-fold: \textbf{(i)} We propose a scalable, fine-grained disentanglement method for categorizing AU by resolvability, based on post-hoc latent space analysis. This is the first uncertainty work to tackle dynamic settings with inconstant input spaces (LI$\Rightarrow$HI) leading to reducible AU. \textbf{(ii)} We put forth a new application of uncertainties in medical contexts, beyond uses for trustworthy AI \cite{oh2024we, bernhardtfailure}, active learning \cite{ma2024breaking, nath2020diminishing}, and semi-supervised learning \cite{wang2021tripled}. In IR imaging of breast tissues, we show multifold gains in downstream segmentation tasks, compared to random re-scanning, by targeting resolvable high-AU pixels. This is the first demonstration of experimental design optimization using uncertainty guidance at a granular level.

\section{Background and Preliminaries}
\label{sec2}
We consider a supervised classification setting, denoting input $\bm{x} \in {X}$ and discrete label $y \in {Y}$ among $C$ classes. Empirical risk minimization (ERM) over training set $\mathcal{D}$ yields a DNN model $w: {X} \rightarrow \Delta_{C-1}$ \textit{s.t.} $w \in \mathcal{H}$, where $\bm{\pi} \in \Delta_{C-1}$ is the probability simplex, and $\mathcal{H}$ is the hypothesis space. Let $w^*$ denote the true data-generating model, or the Bayes predictor with minimal risk over ${X}\times{Y}$. The {total uncertainty} is the error between the predicted $y = \texttt{one\_hot}(\bm{\pi})=\texttt{one\_hot}(w(\bm{x}))$ and the observed true label $y^* \sim \text{Cat}(\bm{\pi^*})=\text{Cat}(w^*(\bm{x}))$. This can be broken down to epistemic (EU) and aleatoric (AU) parts \cite{hullermeier2021aleatoric}—\textbf{EU} arises from $w \neq w^*$, reflecting the learner's ignorance, and stems from model misspecification due to inductive biases ($\mathbb{E}[w] \neq w^*$) and the ERM variance ($\mathbb{V}[w]>0$). On the other hand, \textbf{AU} reflects the intrinsic fuzziness or stochasticity in the data itself, quantified by the Shannon entropy $\mathbb{H}[\bm{\pi^*}]$.

\textbf{Estimations.} AU is estimated by some form of $\mathbb{H}[\bm{\pi}]$, which holds when $\bm{\pi}\approx\bm{\pi^*}$. This is true for low-EU data constrained by ERM, such that $w(\bm{x})\approx w^*(\bm{x})$, $\forall w$, despite $w \neq w^*$. Thus, low EU is a necessity for AU estimations \cite{hullermeier2021aleatoric, wimmer2023quantifying}. EU estimation is more elusive due to its ill-defined nature. Bayesian methods learn the posterior distribution of $w$ \cite{blundell2015weight, maddox2019simple, zhang2019cyclical, gal2016dropout, lakshminarayanan2017simple}, with information-theoretic decomposition of total entropy into conditional entropy (AU) and mutual information (EU) \cite{depeweg2018decomposition, smith2018understanding, kendall2017uncertainties}, but this relies on improper metrics \cite{wimmer2023quantifying} and neglects the EU's bias term \cite{lahlou2023deup}. Bregman decomposition \cite{gruber2023uncertainty} addresses these issues but still struggles to unmix EU/AU \cite{mucsanyi2024benchmarking}, highlighting the limitations of such formulations \cite{ulmer2021know, de2024disentangled, valdenegro2022deeper, mucsanyi2024benchmarking, wimmer2023quantifying, oh2024we}. Additionally, they face scalability issues from multiple DNN Monte Carlo forward passes to perform marginalization. To address this, \textbf{deterministic methods}, such as evidential learning \cite{charpentier2020posterior}, excess risk predictors \cite{lahlou2023deup}, and latent space estimators \cite{lee2018simple, sun2022out, postels2022practicality, mukhoti2023deep}, measure EU through a single DNN pass. Among these, a recent benchmark \cite{mucsanyi2024benchmarking} shows that latent spaces are especially effective at isolating EU by measuring post-hoc\footnote{Some works incorporate regularization like spectral norm for bi-Lipschitzness \cite{mukhoti2023deep} to prevent feature collapse, though studies without it are also successful \cite{lee2018simple, sun2022out}.} ``distance'' to low-EU samples in $\mathcal{D}$. This assumes similar data cluster in DNN latent spaces—a property synonymous with well-learned representations—thereby explaining the strong generalization of latent space EU estimators, irrespective of architecture or training method.

\textbf{Static \textit{vs.} Dynamic.} Most deep learning frameworks operate in a \textit{static} joint space ${X}\times{Y}$, where AU is deemed irreducible (\textit{cf.} EU is reducible via additional data or better choice of $\mathcal{H}$). However, in a \textit{dynamic} LI$\Rightarrow$HI scenario, AU at LI can be resolved by transitioning to a richer, higher-dimensional HI space. However, this shift increases EU, as more complex inputs are inherently harder to learn (``curse of dimensionality''). See \S2.3 in \cite{hullermeier2021aleatoric}.

\textbf{Fine-Grained Disentanglement.} A few studies have explored EU subcategories. For example, \cite{malinin2018predictive, sun2023flagged} divide EU into OOD \textit{vs.} miscalibrated in-distribution cases, while \cite{yang2023full, tian2021exploring} differentiate OOD uncertainty by semantic \textit{vs.} covariate shifts. Our study differs by focusing on finer AU subcategories.

\section{Uncertainty Quantification Method}
\label{sec3}
\textbf{Problem definition.} 
Suppose two domains, LI and HI, have shared label space but distinct DNN models trained on identical tissue samples that differ only in imaging quality. We operate primarily in LI and reserve HI data queries for uncertain LI samples that become certain in HI; our goal is to design a decision function that effectively performs this task.

\subsection{Static Uncertainty Taxonomy}
\label{sec31}
\begin{figure}[t]
\centering
\includegraphics[width=\textwidth]{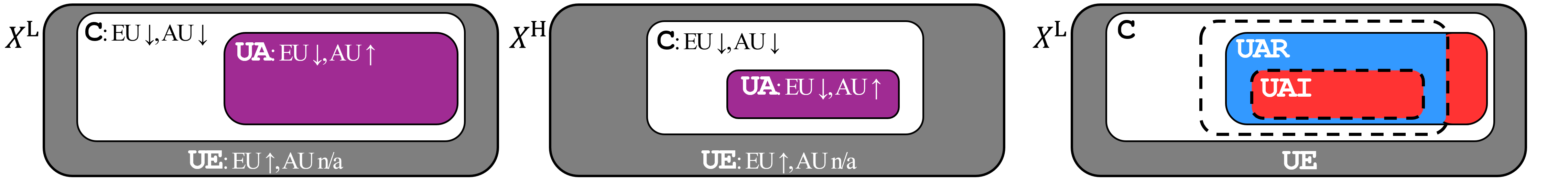}
\caption{\textbf{Visualization of uncertainty categories.} Left to right: static scenarios at LI and HI, and dynamic scenario with LI$\Rightarrow$HI transition (ours).} \label{fig2}
\end{figure}

We begin by considering uncertainty estimates in a static scenario, applying to both LI and HI domains. Let $w = \texttt{softmax} \circ g \circ f$, where $f$ is the feature extractor and $g$ is the final layer such that $\bm{z} = f(\bm{x})$. The generalized formalism for latent space EU and AU estimates is given by:
\small
\begin{equation}
\begin{split}
EU(\bm{x}) \triangleq & \,\,d(\bm{z}; \mathcal{D}_{v}),\\
  AU(\bm{x}) \triangleq &
\begin{cases}
    \mathbb{H}[\bm{\pi}] & \text{if} \quad EU(\bm{x}) < \tau_{{EU}}, \\
    \text{n/a} & \text{otherwise},
  \end{cases}
\end{split}
  \label{eq1}
\end{equation}
\normalsize where $\mathcal{D}_{v} \subseteq \mathcal{D}$ is a training subset (or validation set, if allowed), and $d(\cdot;\mathcal{D}_{v})$ is a distance function \textit{w.r.t.} $\{f(\bm{x_v}) \mid \bm{x_v} \in \mathcal{D}_v\}$, where larger values indicate a higher EU. We define a decision function $\Phi_{\text{St.}}(\cdot)$ with three categories:
\small
\begin{equation}
  \Phi_{\text{St.}}(\bm{x}) \triangleq
\begin{cases}
    \text{Certain: {\texttt{\textbf{C}}}} & \text{if} \quad \bigl(EU(\bm{x}) < \tau_{{EU}}\bigl)\: \land \: \bigl(AU(\bm{x}) < \tau_{AU}\bigl),  \\
    \text{Uncertain} 
    \begin{cases}
        \text{Aleatoric: {\texttt{\textbf{UA}}}} \\
        \text{Epistemic: {\texttt{\textbf{UE}}}} \\
    \end{cases}
    & \begin{array}{l}
        \text{if} \quad \bigl(EU(\bm{x}) < \tau_{{EU}}\bigl)\: \land \: \bigl(AU(\bm{x}) \geqslant \tau_{AU}\bigl),\\
        \text{otherwise} \quad EU(\bm{x}) \geqslant \tau_{{EU}},
    \end{array}
  \end{cases}
    \label{eq2}
\end{equation}
\normalsize with thresholds $\tau_{EU}$ and $\tau_{AU}$ governing category boundaries. While these are, in principle, user-defined hyperparameters, we follow literature-guided protocols to methodologically set them (described later). We schematically delineate $\Phi_{\text{St.}}$ for the LI and HI spaces in Fig. \ref{fig2}. As per \cite{hullermeier2021aleatoric}, EU and AU boundaries tighten in HI, resulting in larger and smaller fractions of \texttt{UE} and \texttt{UA} data, respectively.

\subsection{Dynamic Uncertainty Taxonomy}
\label{sec32}
We now consider the transition from LI to HI, with notations denoted by superscripts L and H, respectively. As evident from Fig. \ref{fig2}, not all \texttt{UA} data at LI are \texttt{C} at HI (\textit{i.e.}, resolvable)—some become \texttt{UE}, while others remain as \texttt{UA}. We seek only the resolvable cases, grouping the rest as irresolvable. Here, we define a new fine-grained uncertainty taxonomy, $\Phi^{\text{Dy.}}(\cdot)$, \textit{w.r.t.} LI data $\bm{x^{\text{L}}}$:
\small
\begin{equation}
  \Phi_{\text{Dy.}}(\bm{x^{\text{L}}}) \triangleq
\begin{cases}
    \text{Cert.: {\texttt{\textbf{C}}}} & \text{if} \quad \Phi^{\text{L}}_{\text{St.}}(\bm{x^{\text{L}}})={\texttt{\textbf{C}}}, \\
    \text{Uncert.} 
    \begin{cases}
        \text{Al.} 
        \begin{cases}
            \text{Res.:  {\texttt{\textbf{UAR}}}} \\
            \text{Irres.: {\texttt{\textbf{UAI}}}} \\
        \end{cases}\\
        \text{Ep.: {\texttt{\textbf{UE}}}}
    \end{cases}
    & \begin{array}{l}
        \text{if} \quad \bigl(\Phi_{\text{St.}}^{\text{L}}(\bm{x^{\text{L}}})={\texttt{\textbf{UA}}}\bigl) \land \bigl(\Phi_{\text{St.}}^{\text{H}}(\bm{x^{\text{H}}})=\texttt{\textbf{C}}\bigl),\\
        \text{if} \quad \bigl(\Phi_{\text{St.}}^{\text{L}}(\bm{x^{\text{L}}})={\texttt{\textbf{UA}}}\bigl) \land \bigl(\Phi_{\text{St.}}^{\text{H}}(\bm{x^{\text{H}}})\neq\texttt{\textbf{C}}\bigl),\\
        \text{otherwise} \quad \Phi_{\text{St.}}^{\text{L}}(\bm{x^{\text{L}}})={\texttt{\textbf{UE}}}.
    \end{array}
  \end{cases}
    \label{eq3}
\end{equation}
\normalsize During test-time, explicitly using this framework is infeasible as it requires foreknowledge of HI data from $\Phi_{\text{St.}}^{\text{H}}(\bm{x^{\text{H}}})$. Instead, following similar approach used for EU estimation, we approximate the decision criteria of classes $\forall \phi \in \{\texttt{UAR}, \texttt{UAI}\}$ by leveraging latent space distances in the LI domain, as follows:
\small
\begin{equation}
\begin{split}
\Phi_{\text{Dy.}}(\bm{x^{\text{L}}})\approx\phi \quad \text{if} \quad \text{arg min}_{\phi} d(\bm{x^{\text{L}}};\mathcal{D}_{v}^{\phi})=\phi, \\
s.t. \quad \mathcal{D}_{v}^{\phi} \triangleq \{\bm{x^{\text{L}}_v} \mid \Phi_{\text{Dy.}}(\bm{x^{\text{L}}_v}) = \phi, \forall \bm{x_v} \in \mathcal{D}_{v}\}.
\end{split}
\label{eq4}
\end{equation}
\normalsize That is, we assign \texttt{UAR} \textit{vs.} \texttt{UAI} by evaluating the proximity to respective LI prototypes within $\mathcal{D}_{v}$, for which we can compute Eq. \ref{eq3}, since both their LI and HI data are available. The surrogate Eq. \ref{eq4} obviates the need for HI test data, thereby enabling blind uncertainty inference directly in LI, while operating post-hoc with minimal overhead—an advantage in digital pathology, where data can sum up to TBs. However, since we aim to emulate Eq. \ref{eq3}, its performance is upper-bounded by it. This hinges on static uncertainties at LI and HI (Eqs. \ref{eq1}-\ref{eq2}), and while we opt for latent space formulations due to their state-of-the-art performance \cite{mucsanyi2024benchmarking}, it is not restricted to this backbone. Unless specified otherwise, we use the popular Mahalanobis distance (MD) \cite{lee2018simple, podolskiy2021revisiting, dua2023task} for all $d(\cdot;\cdot)$ metrics, measuring proximity to nearest Gaussian class centroid.

\subsection{Application in Adaptive Imaging}
\label{sec33}
Selecting class-\texttt{UAR} LI data for HI query yields the final adaptive prediction:
\small
\begin{equation}
\bm{\pi^{\text{A}}} = \underbrace{\mathbbm{1}[\Phi_{\text{Dy.}}(\bm{x^{\text{L}}})\neq\texttt{UAR}]\cdot w^{\text{L}}(\bm{x^{\text{L}}})}_{\text{Existing default LI data}} + \underbrace{\mathbbm{1}[\Phi_{\text{Dy.}}(\bm{x^{\text{L}}})=\texttt{UAR}]\cdot w^{\text{H}}(\bm{x^{\text{H}}})}_{\text{Newly queried HI data}}.
\label{eq5}
\end{equation}
\normalsize Note that not all data are certain—only \texttt{C} and, for well-approximated Eq. \ref{eq4}, \texttt{UAR}. Let $T^\text{H}$ and $T^\text{L}$ represent the imaging complexity in HI and LI, respectively. The cost of Eq. \ref{eq5} is then $T^{\text{A}}=T^{\text{L}} + P_{\bm{x}\sim X}(\Phi_{\text{Dy.}}(\bm{x^{\text{L}}})=\texttt{UAR})\cdot T^\text{H}$. Still, this may be prohibitive with larger \texttt{UAR} ratios or tighter budgets. In such circumstances, we chose by ascending values of $d(\bm{x^{\text{L}}}; \mathcal{D}_v^{\texttt{UAR}})$, stopping once the allowable query limit is reached; this prioritizes data with higher ``confidence'' of belonging to \texttt{UAR}.

\section{Experiments}
\label{sec4}

\subsection{Experimental setting}
\label{sec41}
Here, we describe the conducted experiments to showcase our approach.

\textbf{Dataset.} We use the IR dataset of a breast cancer tissue microarray consisting of 101 cores from 47 patients, provided by \cite{mittal2018simultaneous}. We select 10 IR spectral bands at 1.1$\upmu$m/pixel resolution as HI, simulating the LI counterparts by reducing the number of bands to 4 and spatially downsampling by 10×—a 250× difference in imaging complexity. The downstream task is 6-class segmentation by epithelial and stromal types, with an 80-20\% training-test split. Since segmentation is dense per-pixel classification, all notations and concepts transfer.

\textbf{Implementations.} We train four models\footnote{Attention-Unet architecture \cite{oktayattention}, trained for $\sim$5K iterations using a batch size of 16, focal loss \cite{lin2017focal}, pixel dropout, and AdamW \cite{loshchilov2017decoupled} with an initial learning rate of 1e-4.} per domain (with differing seeds), resulting in 16 adaptive imaging outcomes per pair. As the task is segmentation, we generate dense uncertainty maps from the final feature map, with adaptive imaging occurring at the pixel level. To reduce uncertainty inference latency, we downscale these maps by 4×. We set implementation hyperparameters as so: For $\mathcal{D}_v$, we draw $\sim$6K training pixels stratified by class and core, following the smallest sampling ratios in \cite{sun2022out, lee2018simple, kim2024neural} for stringent conditions. $\tau_{EU}$ is set to achieve a 95\% true positive rate in the low-EU training set, as per routine OOD detection evaluation protocols \cite{hendrycks2022baseline, chan2021entropy, lee2018simple, sun2022out}. $\tau_{AU}$ is calibrated to maximize the number of accurate \texttt{C} data, matching the $P(accurate, certain)$ metric \cite{mukhoti2018evaluating}. We also tested transformer architectures \cite{cao2022swin} and hyperparameter configurations, reaching the same conclusion (omitted due to space constraints).

\begin{figure}[t]
\centering
\includegraphics[width=0.97\textwidth]{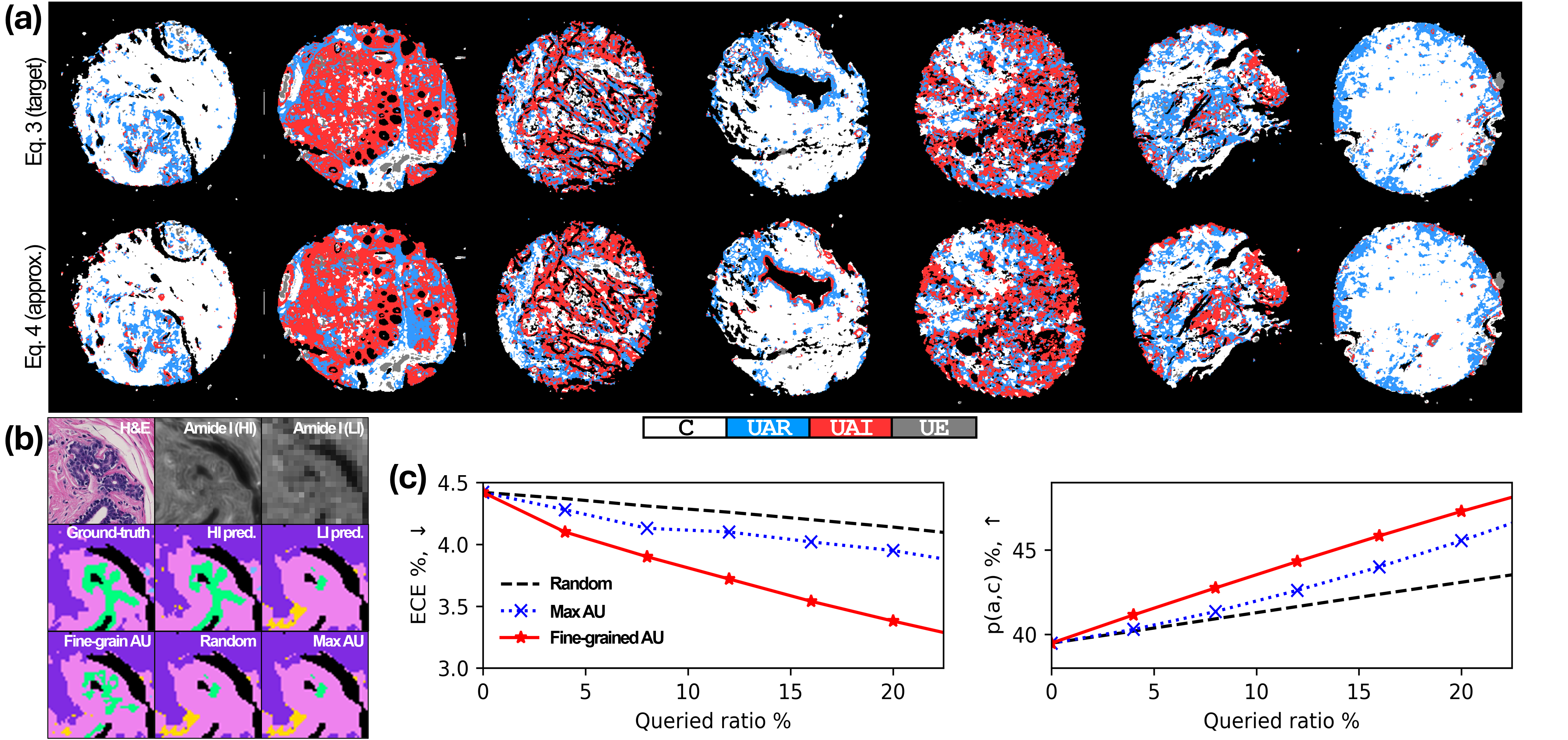}
\caption{\textbf{Results.} \textbf{(a)} Comparison of $\Phi_{\text{Dy.}}$: Eq. \ref{eq3} (top row) \textit{vs.} Eq. \ref{eq4} (bottom row). \textbf{(b)} Example segmentation results via adaptive imaging using different query strategies. \textbf{(c)} Segmentation performance metrics under tighter query budgets.} \label{fig3}
\end{figure}

\textbf{Baselines.} We compare against two query baselines: \textit{random}, and \textit{maximum AU} among samples where $\Phi_{\text{St.}}^{\text{L}}(\bm{x^{\text{L}}})=\texttt{UA}$. These baselines serve to prove our core claims: the random baseline asserts the value of prioritizing high-AU pixels, while the monolithic max-AU baseline tests whether finer subcategorization yields additional benefit. Although alternative (non-uncertainty-based) sampling strategies exist, we do not consider them here due to space constraints.

\subsection{Results \& Discussion}
\label{sec42}
Next, we present our experimental findings, with in-depth analysis. 

\textbf{Fine-grained AU disentanglement via latent spaces is feasible.} As a sanity check, we first examine whether Eq. \ref{eq3} using latent spaces can approximate Eq. \ref{eq4}. We compare the uncertainty class maps for selected test cores in Fig. \ref{fig3}-(a), observing a strong agreement between the two, with a $\sim$60\% mean F1 score for \texttt{UAR} and \texttt{UAI}. This affirms the feasibility of our approach: \textit{latent spaces can not only capture EU extent but also disentangle fine-grained AU subtypes}. We hypothesize this is due to \texttt{UAR} and \texttt{UAI} being low-EU categories, resulting in robust, accurate latent representations—facilitating separation via a suitable distance function $d(\cdot;\cdot)$. Notably, we made no assumptions about the LI/HI domains, rendering our approach versatile and extendable to any uncertainty taxonomy, {provided their uncertainty class prototypes are accessible.} 

\textbf{Querying by AU resolvability is efficacious.} In Fig. \ref{fig3}-(b), we show an example of a segmented region with a fixed adaptive query budget, where our method approaches the ground-truth most proficiently. We quantify three segmentation metrics in Tab. \ref{tab1}, with arrows indicating the desired direction: F1 ($\uparrow$), Expected Calibration Error (ECE, $\downarrow$) \cite{kumar2019verified}, and the above $P(a,c)$ ($\uparrow$) \cite{mukhoti2018evaluating}. We evaluate two budget scenarios—\textit{(i) Unconstrained}, allowing an arbitrary number of queries (Eq. \ref{eq5}). In our dataset, the proportion of \texttt{UAR} was $\sim$20\%, corresponding to a cost $T^{\text{A}}=50$ when denoting $T^{\text{L}}$ and $T^{\text{H}}$ as 1 and 250 (a.u.). \textit{(ii) Constrained}, with tighter budgets. We compute the metrics spanning $1<T^{\text{A}}<50$ (Fig. \ref{fig3}-(c)) and report the area-under-curve (AUC), denoted by $\int$. The results show that our method achieves superior ECE and $P(a,c)$, with gains over LI of 1.14/8.21\% and 0.92/6.35\% for $T^{\text{A}}=50$ and $1<T^{\text{A}}<50$, respectively. Although these improvements may appear modest due to the upper bound imposed by HI performance, they constitute substantial relative gains—exceeding the random and max-AU baselines by \textbf{2.1–5.7×} and \textbf{1.4–3.2×}, respectively. While the max-AU marginally outperforms in F1, its inferior ECE and $P(a,c)$ indicate a tendency to select high-AU samples that, while becoming more accurate, remain poorly calibrated or uncertain (i.e., do not cross the $\tau_{AU}$ threshold); arguably, re-imaging such samples are less valuable.

\begin{table*}[t]
\small
\centering
\caption{\textbf{Comparison of query methods.} The best method is in bold. The LI/HI performance metrics are as follows—F1\%: 53\textsubscript{±0.8}/60.5\textsubscript{±1.3}, ECE\%: 4.3\textsubscript{±0.4}/2.1\textsubscript{±0.8}, and $P(a,c)$\%: 41.6\textsubscript{±2}/54.6\textsubscript{±2.5}.}
\resizebox{1\textwidth}{!}{%
\begin{tabular}{l|ccc|ccc}
\toprule
& \multicolumn{3}{c}{Unconstrained, $T^{\text{A}}\approx50$} & \multicolumn{3}{c}{Tighter constraints, $1<T^{\text{A}}<50$} \\
\cmidrule(lr){2-4} \cmidrule(lr){5-7}
Method & F1\%,$\uparrow$ & ECE\%,$\downarrow$ & $P(a,c)$\%,$\uparrow$ & $\int$F1\%,$\uparrow$ & $\int$ECE\%,$\downarrow$ & $\int$$P(a,c)$\%,$\uparrow$ \\
\midrule

Random & 54.46\scriptsize{±0.55} & 3.77\scriptsize{±0.39} & 44.04\scriptsize{±1.38} & 53.70\scriptsize{±0.58} & 4.06\scriptsize{±0.35} & 42.75\scriptsize{±1.54} \\

Max AU & \textbf{55.80\scriptsize{±0.54}} & 3.51\scriptsize{±0.46} & 46.51\scriptsize{±1.59} & \textbf{54.45\scriptsize{±0.49}} & 3.89\scriptsize{±0.36} & 43.59\scriptsize{±1.64} \\

Fine-grained AU & 54.90\scriptsize{±0.81} & \textbf{3.16\scriptsize{±0.53}} & \textbf{49.84\scriptsize{±1.52}} & 53.97\scriptsize{±0.77} & \textbf{3.38\scriptsize{±0.46}} & \textbf{47.98\scriptsize{±1.75}} \\


\bottomrule
\end{tabular}
}
\label{tab1}
\end{table*}

\textbf{Assumption violations cause failures.} Recall that our approach hinges on two core premises—(i) the strong base performance of static LI and HI uncertainties, and (ii) the well-approximation of Eq. \ref{eq4}. A failure in either premise undermines our method. To demonstrate this, we conduct ablation experiments: (i) We replace the latent space static uncertainties with an inferior alternative—the information-theoretic EU/AU estimates, which are popular in medical fields \cite{huang2024review}, despite known limitations \cite{ulmer2021know, valdenegro2022deeper, wimmer2023quantifying, mucsanyi2024benchmarking, oh2024we, de2024disentangled}. Specifically, we adopt Deep Ensemble (DE) \cite{lakshminarayanan2017simple}, where the four DNNs per domain are treated as constituents. We use the same MD-based Eq. \ref{eq4}, but with averaged distances across DE members. (ii) We compute MD using only the feature's 1\textsuperscript{st} principal component (PC) in Eq. \ref{eq4}, anticipating diminished performance due to the loss of information; this modification is applied solely to Eq. \ref{eq4}, leaving the static latent space uncertainties (Eqs. \ref{eq1}-\ref{eq2}) unaltered.

\begin{table*}[t]
\small
\centering
\caption{\textbf{Other configurations.} First row is the original version, followed by ablations and design variations. For ablations, we highlight in red if there is a significant drop in performance. No ±std. is reported for the DE ablation, as we aggregate all models. We also report uncertainty inference latency (s/mm\textsuperscript{2}, $\downarrow$) on an i9-9900KF CPU in \textcolor{blue}{parentheses} for the original method and design variations.
}
\resizebox{1\textwidth}{!}{%
\begin{tabular}{l|ccc|ccc}
\toprule
Method & $\uptau$,$\approx0$ & AUCC,$\downarrow$ & Apx.F1\%,$\uparrow$ & $\int$F1\%,$\uparrow$ & $\int$ECE\%,$\downarrow$ & $\int$$P(a,c)$\%,$\uparrow$ \\
\midrule
Original; MD w/o DR \textcolor{blue}{(0.97)} & -0.94\scriptsize{±5.68} & 2.46\scriptsize{±0.70} & 60.39\scriptsize{±1.27} & 53.97\scriptsize{±0.77} & {3.38\scriptsize{±0.46}} & {47.98\scriptsize{±1.75}} \\

\midrule

Abl. (i): $\Phi_{\text{St.}}$ w/ DE & \cellcolor{red!30}30.88 & \cellcolor{red!30}6.58 & 65.38 & 55.58 & \cellcolor{red!30}6.33 & \cellcolor{red!30}44.95 \\

Abl. (ii): $d$ in Eq. \ref{eq4} w/ PC1 & -0.94\scriptsize{±5.68} & 2.46\scriptsize{±0.70} & \cellcolor{red!30}49.08\scriptsize{±0.52} & 53.78\scriptsize{±0.61} & \cellcolor{red!30}3.88\scriptsize{±0.34} & \cellcolor{red!30}42.99\scriptsize{±1.59} \\

\midrule

KNN w/o DR \textcolor{blue}{(3.11)} & 7.92\scriptsize{±5.16} & 2.88\scriptsize{±1.25} & 60.70\scriptsize{±1.79} & 53.70\scriptsize{±0.77} & 3.22\scriptsize{±0.43} & 48.03\scriptsize{±1.47} \\

MD w/ DR \textcolor{blue}{(0.82)} & -9.24\scriptsize{±11.2} & 3.14\scriptsize{±1.18} & 59.65\scriptsize{±2.58} & 53.85\scriptsize{±0.77} & 3.33\scriptsize{±0.43} & 47.24\scriptsize{±1.26} \\


\bottomrule
\end{tabular}
}
\label{tab2}
\end{table*}

In Tab. \ref{tab2}, we report the segmentation metrics for these ablations. Since the exact number of \texttt{UAR} data slightly varied between methods, we report only the constrained budget scenario to ensure a fair comparison. Additionally, we evaluate the satisfaction of hypotheses (i)-(ii) as follows: (i) Following \cite{mucsanyi2024benchmarking}, we adopt two metrics to assess static uncertainties, averaged across LI and HI—rank correlation (Kendall's $\uptau$, where $\approx$0 is desired) between EU and AU to measure the degree of entanglement, and the AUC of Calibration-Coverage (AUCC, $\downarrow$), where we abstain from predictions by decreasing EU and compute the ECE of the remaining. (ii) As previously done, we report the mean F1 ($\uparrow$) score of \texttt{UAR} and \texttt{UAI} approximations. As expected, targeted ablations cause a drop in the corresponding metric(s). Consequently, we observe decreased segmentation performance, highlighting the importance of maintaining assumptions.

\textbf{Robustness to design variations.} Alternative distances for $d(\cdot;\cdot)$, such as KNN \cite{sun2022out}, are also viable, while some works apply dimensionality reduction (DR) to mitigate the curse of dimensionality \cite{ghosal2024overcome, li2024characterizing}. We also explore these options in Tab. \ref{tab2}. Setting $k=100$ for KNN, we observed slightly better performance due to its nonparametric nature. However, the gap is small, leading us to choose the more efficient MD with an uncertainty latency of 0.97s/mm\textsuperscript{2}. For a typical whole-slide tissue image covering a 1cm\textsuperscript{2} area, this translates to 1-2 minutes, which can be reduced by further downsampling the uncertainty map, making it highly practical. Note, however, that we expect KNN to significantly outperform MD for heterogeneous or long-tailed latent distributions. For DR, we tried multiple manifold learning algorithms like Isomap, LLE, and Parametric UMAP \cite{sainburg2021parametric}, reporting only the latter, which performed the best. However, it provided no significant improvement, likely because our final feature vector dimension is only 64, typical in segmentation DNNs. As such, the benefit of DR is weaker, potentially offset by information loss. However, we anticipate that it will be more crucial in image classification extensions of our task, where feature dimensionality is much larger (512+). Nonetheless, all variants performed well, supporting the robustness of our approach.

\subsection{Concluding Comments}
\label{sec43}
Having shown proof-of-concept, we conclude with several avenues for future research. {(i) Even finer disentanglement:} In this work, LI degradation occurred along two axes—chemical (via band reduction) and spatial (via sparse sampling)—which we abstracted into a binary \texttt{UAR} vs. \texttt{UAI} distinction. However, not all \texttt{UAR} data are equal; some can be more cheaply resolved by correcting just one dimension. This suggests that adaptive imaging could be further optimized through even finer (2+) AU disentanglement. {(ii) Algorithmic improvements:} One direction is to enhance static uncertainty estimates, e.g., through graph-based OOD/failure detectors \cite{goodge2022lunar, kim2024neural} for EU and heteroscedastic DNNs \cite{colliermassively, kendall2017uncertainties} for AU. Another question is how to optimize $d(\cdot;\cdot)$ for fine-grained separation tasks. (iii) Variable LI/HI cost differential: While we assume a fixed cost ratio, reducing LI cost could enable larger query budgets and potentially improve net re-imaging quality. Although our approach assumes no specific cost structure and is thus broadly applicable, further analysis is needed to determine optimal strategies under varying cost regimes. {(iv) Large-scale validations:} Due to the lack of publicly available label-free histology datasets, we limited our study to just one. We plan to pursue further validation upon the release of more datasets.


    

\begin{credits}
\subsubsection{\ackname} Research reported in this publication was supported by the National Institutes of Health under award numbers R01EB009745, R01CA260830, and R21CA263147. The content is solely the responsibility of the authors and does not necessarily represent the official views of the National Institutes of Health. This work was also supported by NSF grant 2243257, the National Science Foundation Science and Technology Center for Quantitative Cell Biology. Rohit Bhargava is a CZ Biohub Investigator.
 
\subsubsection{\discintname}
The authors declare no conflict of interest.
\end{credits}

%
%
%
%
\bibliographystyle{splncs04}
\bibliography{bib}

\end{document}